# Wide-Residual-Inception Networks for Real-time Object Detection

Youngwan Lee[1], Byeonghak Yim[2], Huieun Kim[2], Eunsoo Park[2], Taekang Woo[2], Xuenan Cui[2], Hakil Kim[2]*

*Abstract*— Since convolutional neural network (CNN) models emerged, several tasks in computer vision have actively deployed CNN models for feature extraction. However, the conventional CNN models have a high computational cost and require high memory capacity, which is impractical and unaffordable for commercial applications such as real-time on-road object detection on embedded boards or mobile platforms. To tackle this limitation of CNN models, this paper proposes a wide-residual-inception (WR-Inception) network, which constructs the architecture based on a residual inception unit that captures objects of various sizes on the same feature map, as well as shallower and wider layers, compared to state-of-the-art networks like ResNets. To verify the proposed networks, this paper conducted two experiments; one is a classification task on CIFAR-10/100 and the other is an on-road object detection task using a Single-Shot Multi-box Detector (SSD) on the KITTI dataset. WR-Inception achieves comparable accuracy on CIFAR-10/100, with test errors at 4.82% and 23.12%, respectively, which outperforms 164-layer Pre-ResNets. In addition, the detection experiments demonstrate that the WR-Inception–based SSD outperforms ResNet-101–based SSD on KITTI. Besides, WR-Inception–based SSD achieves 16 frames per seconds, which is 3.85 times faster than ResNet-101-based SSD. We could expect WR-Inception to be used for real application systems.

## I. Introduction

Due to the recent active studies and achievements regarding artificial intelligence (AI), AI technologies based on deep neural networks (DNNs) are actively utilized for many different fields in society, and the current trend is that they are required in even more areas. In particular, the emergence of convolutional neural networks (CNNs, or ConvNets) [1], [2] in computer vision has been replacing traditional computer vision technology. The CNN models not only enhance the accuracy of image classification [3]–[9] but they are also used as the generic feature extractor [10]–[12] in the fields of object detection [13]–[20], semantic segmentation [21]–[23], and depth estimation [24].

However, this CNN technology has a high computational cost and requires a lot of memory, and in order to train and deploy it, a high-specification hardware system is necessary. A system to be put in an advanced driver assistance system (ADAS), or self-driving cars, requires a real-time processing capability even in an embedded board, which has relatively limited computing power. An embedded board has many limitations, compared to a desktop PC, in terms of computing power, power consumption, memory, and other properties, and so there are restrictions on applying DNN-based algorithms and systems that require extensive computations. Therefore, studies into optimization of CNN technology to overcome such limits are needed.

Therefore, in order to tackle these difficulties, this research proposes a wide-residual-inception (WR-Inception) network, which shows similar performance to the latest deep neural network model but with less memory weight and fewer computations. As a method to solve the issue of gradient vanishing, this study applies residual connections [5] and proposes a residual inception unit that can see various receptive fields.

The contributions of this study are that it

- proposes a model for a WR-Inception network that requires less memory and fewer computations but shows better performance

- achieves better performance than state-of-the-art network models when applying the model to the feature extraction network of an object detector, and

- is capable of real-time processing of a DNN-based object detector in an embedded board

The contents of this paper are as follows. We introduce the trends in related research projects in Section II, and cover the proposed WR-Inception network in Section III. Section IV deals with the WR-Inception network's transfer learning to an object detector. Section V shows image classification, the object-detecting experiment, and the resulting analysis, and Chapter 6 offers conclusions.

## II. Related works

Since the advent of AlexNet [2] with eight layers, the models have had a tendency to increase the depth of the network for the model's capabilities. For example, the VGG network [4] has 16 convolutional layers and three fully connected layers and ranked second in the ImageNet Large Scale Visual Recognition Challenge (ILSVRC) 2014, and GoogleNet [3] consists of 21 convolutional layers and one fully connected layer, and ranked first in ILSVRC 2014. However, increasing the depth of networks causes the vanishing gradient problem as well as the over-fitting problem. To prevent vanishing gradients, many methods have been introduced, such as MSR initialization [25], various activation function ReLU, ELU [37], PreLU [25], and PELU [26], and Batch normalization [27].

Meanwhile, ResNets proposed skip connection (identity mapping) to deal with this degradation problem by propagating the information to deeper layers without vanishing, which enables increases of up to thousands of layers, and helped to win five major image recognition tasks in ILSVRC 2015 and Microsoft Common Objects in Context (MS-COCO) 2015 competitions.

*corresponding author

[1] Youngwan Lee is with Electronics and Telecommunications Research Institute, Daejeon, Korea. email: yw.lee@etri.re.kr

[2] HuiEun Kim(M.S.), Byeonghak Yim(M.S.), Taekang Woo(M.S.), Eunsoo park(Ph.D candidate), Xuenan Cui (Professor), and Hakil Kim(Professor) is with the Information and Communication Engineering Department, Inha University, Incheon, 22212, Korea. email: {hekim | bhy516 | tkwoo | espark}@inha.edu,{ xncui | hikim }@inha.ac.kr

The main idea of residual networks is identity skip-connections, which skip blocks of convolutional layers to help gradients to bypass the weight layers, forming a shortcut residual block (residual unit). Residual blocks are shown in Fig. 1, where each residual block can be represented as follows.

$$x_{l+1} = x_l + F(x_l, W_l) \qquad (1)$$

However, one shortcoming of deep residual networks is that increasing the depth of the network requires a high computational cost and a large memory capacity, which is impractical and not economic for commercial products that have limited hardware resources.

In addition, the research of Veit et al. [28] demonstrated that ResNets actually behave like ensembles of relatively shallow networks, not as single deep networks, and they do not resolve the vanishing gradient problem by preserving gradient flow through the entire network. Rather, they avoid the problem by ensembling short networks. They experimentally proved that most gradients in ResNet-101 come from an ensemble of very short networks, i.e., only 10 to 34 layers deep. Regarding these perspectives, this paper tries to find the proper depth of networks and architectures for practical uses.

### III. WIDE-RESIDUAL-INCEPTION NETWORKS

#### A. Factors to Consider in Neural Network Modeling—Width vs. Depth

He et al. [29] experimentally claimed that a network has three factors that are most important when constructing a network architecture: depth, width, and filter size. When the time complexity was fixed and the trade-offs between depth vs. width, depth vs. filter size, and width vs. filter size were tested, the results of the experiments prioritizing depth showed the highest performance.

On the other hand, a paper about wide-residual networks [9] proved that while maintaining the shortcut connection of ResNets, a wide and shallow network model (not a thin and deep one, like a ResNet) could outperform ResNets.

Therefore, this study proposes a network optimized for an embedded environment by applying the two claims experimentally. We proceeded with our network design from the perspective of a "small" network unit (a micro-architecture) and the whole network (the macro-architecture) that is composed of such small units.

#### B. Micro-Architecture

- Basic residual (3x3,3x3):

  The most basic unit places two simplest 3x3 convolutional layers consecutively and connects them with a shortcut connection.

- Bottleneck (1x1, 3x3, 1x1):

  The unit places a 1x1 convolutional layer to reduce the dimension of feature maps, stacks 3x3 and 1x1, subsequently, and restores the dimension in the last 1x1 convolution.

- Inception:

  The network unit contains different types of convolutional layers at the same level, i.e., 1x1, 3x3, and 5x5 convolutional layers are included in the same feature level, which captures objects at various scales proposed by GoogleNet.

#### C. Residual Inception Unit

Fig. 1 (c) shows the residual-inception within the proposed network. This adds a shortcut connection to the inception module, and rather than a 1x1 convolution to each branch, it merges each 1x1 convolutional layer before 3x3 convolutional layers, and subsequently, consists of two consecutive 3x3 convolutional layers that have the same operational result as one 5x5 convolutional layer and one 3x3 convolutional layer from a single 1x1 convolutional layer. Through a concatenation operation, it made the three branches into a single tensor, and expanded the feature map space. From that, as shown in Figure 3, it could extract various receptive fields with different scales from one feature map. As a result, from the object detection task, we could obtain an effect to simultaneously extract different-sized objects at the same level of the feature map stage, achieving the enhanced object detection rate.

#### D. Macro-Architecture

This paper proposes a wide-residual-inception (WR-Inception) network where the aforementioned residual-inception unit is applied to a wide-residual network. In order to verify the effect of the residual-inception module, as shown in Table 1, we set all the networks the same, but replace one residual unit of WRN-16-4 with one residual-inception unit in the conv3_x stage.

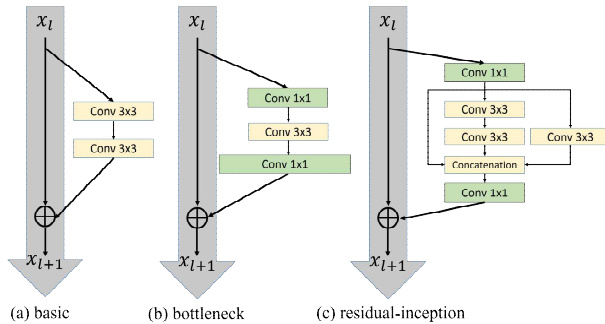

Figure. 1. various residual units

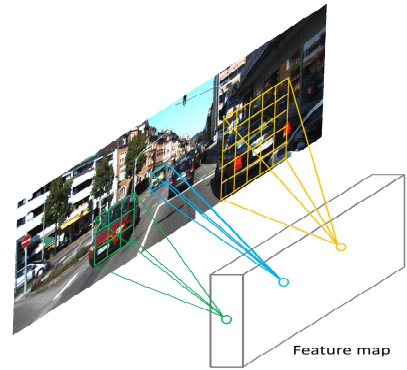

Figure. 2. The effect of various receptive fields

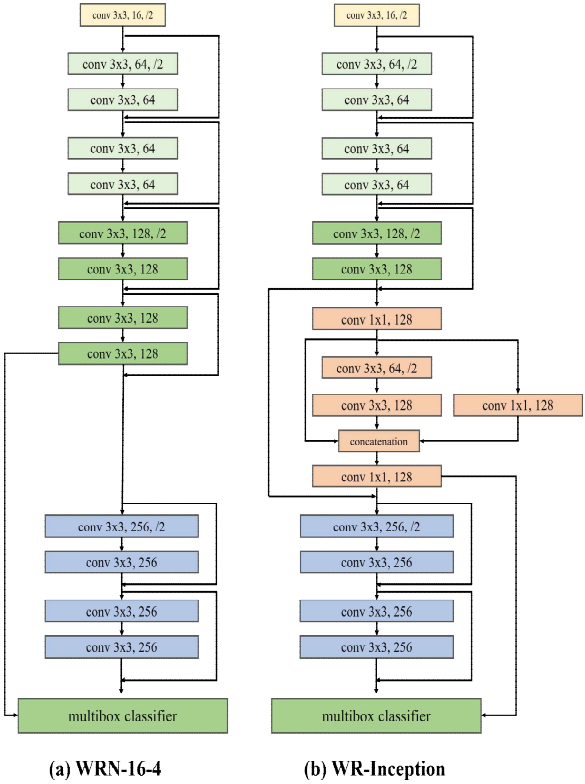

Figure. 3. Comparison of network architectures

$(128×3^2×128)×2$

$128×1^2×128 + 128×3^2×64 + 128×3^2×64 + 64×3^2×128 + 320×1^2×128$

At conv3_x in Table 1, when changing a residual unit of the WRN-16-4 into a residual-inception unit, we compose the residual inception unit to have the same theoretical computational time complexity as a residual unit consisting of consecutive 3x3 convolutional layers with an input dimension of 128 and convolution filters of 128.

One can design WR-Inception networks in different versions by setting the WR-Inception network as the baseline and adjusting the number of convolutional filters at the conv1, conv2_x, conv3_x, and conv4_x stages while considering desirable performance and processing time. The WR-Inception-l2 version in Table 1 is a model with higher performance than other models from doubling the number of convolutional filters at the conv3_x stage while maintaining the real-time processing speed in the TX1 embedded board. Fig. 3 shows the WRN-16-4 network and WR-Inception network.

## IV. TRANSFER LEARNING FOR OBJECT DETECTION

### A. Outline of Transfer Learning

One of the most important properties of ConvNets is that they extract good feature representations. In particular, a ConvNet trained by the ImageNet dataset with 1000 categories and more than 1.2 million images surpasses the recognition rate of humans and extracts the most generic visual feature representations. Accordingly, in many areas of computer vision, the number of cases using the ConvNet model, which plays the role of a well-trained feature extractor as the pre-trained model, is increasing [30], [31].

Fig. 4 is a flow chart of transfer learning. In order to train a ConvNet at the beginning, it sets the initial value using MSR initialization [25] and trains on CIFAR-10/100 or the ImageNet dataset (most frequently used in image classification) for the source data. Then, the "source ConvNet" trained by the source data, referred to as the pre-trained model, is used as the initial weight value of the target task.

After weights are initialized by using the source ConvNet, the whole network is trained (fine-tuned) in accordance with the target task, e.g., object detection or segmentation by the target data of the task to update the weight.

There is a case where all the weights are updated, but since a low-level layer extracts relatively general properties (line, edge, etc.), it may not be necessary to update weights at all. If so, we "freeze" the weights (a metaphor for preventing weights from getting updated).

TABLE 1. CONFIGURATION OF NETWORKS

| Model | WRN-16-4 | | WR-Inception | | WR-Inception-l2 | |
|---|---|---|---|---|---|---|
| conv1 | 3x3,16 | | 3x3,16 | | 3x3,**64** | |
| conv2_x(k) | 3x3,64 | x2 | 3x3,64 | x2 | 3x3,64 | x2 |
| | 3x3,64 | | 3x3,64 | | 3x3,64 | |
| conv3_x(l) | 3x3,128, s/2 | x2 | 3x3,128, s/2 | x1 | 3x3,**256** | x1 |
| | 3x3,128 | | 3x3,128 | | 3x3,**256** | |
| inception(l) | - | | 1x1,128, s/2 | | 1x1,**256**, s/2 | |
| | | | 3x3,64 | 3x3,64 | 3x3,**256** | 3x3,**128** |
| | | | - | 3x3,128 | - | 3x3,**256** |
| | | | 1x1,128 | | 1x1,**256** | x1 |
| conv4_x(m) | 3x3,256, s/2 | x2 | 3x3,256, s/2 | x2 | 3x3,256, s/2 | x2 |
| | 3x3,256 | | 3x3,256 | | 3x3,256 | |

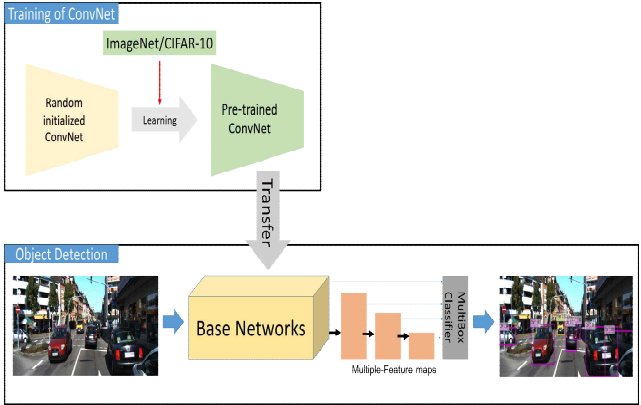
Figure 4. Overview of transfer learning for object detection

## B. Single-Shot Multi-box Detector

As seen in Fig. 4, a Single-Shot Multi-box Detector (SSD)[19] network is largely composed of a base network (feature extraction network) and a classifier network. It uses the best feature map as a result of continuous extraction from the base network, leading to object classification through the softmax cross-entropy loss simultaneously and localization through the bounding box regression using smooth L1 loss in the classifier network.

While general CNN-based object detection algorithms use a single feature map to extract objects, SSD has the advantage of extracting different-sized objects by choosing feature maps at different scales.

The original SSD chooses a feature map by using VGG (which is widely used because of its simple network structure) as the base network, but the network has a problem in that it takes up about 80% of the whole processing time. In order to overcome the problem, this paper replaces VGG with the proposed WR-Inception network and improves processing time, performance, and memory use.

## V. EXPERIMENTAL RESULTS

Our research conducted largely two types of experiment:

- ✓ verification of the proposed WR-Inception network model on the CIFAR-10/100 dataset [32]
- ✓ application of the proposed network to an object detector as the feature extraction network (transfer learning) on the KITTI dataset [33]

### A. Verification of the Network Model

In order to verify the performance of the proposed network model, we used the CIFAR-10/100 dataset, which is composed of a training set of 50,000 images and a test set of 10,000 images sized 32x32. CIFAR-10 has 10 categories, and CIFAR-100 has 100 categories.

For the performance comparison, we set ResNet-164 and WRN-16-4 as the baseline comparison group and trained them in the same way that Zagoruyko[9] did. We used stochastic gradient descent with Nesterov momentum as the weight update method (0.9 for the momentum, 0.005 for the weight decay, and 128 as the mini-batch size), equally distributed the batch to two NVIDIA 1080 GPUs (64 images each), and trained them under the multi-GPU environment. Starting from a learning rate of 0.1, we reduced it to 0.02 at epochs of 60, 120, and 160, and trained 200 epochs in total.

TABLE 2. TEST ERROR (%) ON CIFAR-10/100 BY DIFFERENT NETWORKS

| Network models | Depth | # of Params | CIFAR-10 | CIFAR-100 |
|---|---|---|---|---|
| NIN[35] | | | 8.81 | 35.67 |
| DSN[36] | | | 8.22 | 34.57 |
| FitNet[37] | | | 8.39 | 35.04 |
| Highway[38] | | | 7.72 | 32.39 |
| ELU[39] | | | 6.55 | 24.28 |
| original-ResNet[5] | 110 | 1.7M | 6.43 | 25.16 |
| | 1202 | 10.2M | 7.93 | 27.82 |
| stoc-depth[40] | 110 | 1.7M | 5.23 | 24.58 |
| | 1202 | 10.2M | 4.91 | - |
| pre-act-ResNet[8] | 110 | 1.7M | 6.37 | - |
| | 164 | 1.7M | 5.46 | 24.33 |
| WRN-16-4[9] | 16 | 2.8M | 5.37 | 24.53 |
| **WR-Inception** | 16 | **2.7M** | **5.04** | **24.16** |
| **WR-Inception-l2** | 16 | **4.8M** | **4.82** | **23.12** |

Table 2 is a comparison of the test errors of the CIFAR-10/100 classification. One can note that our proposed model had 1.33% and 2.83% lower error rates than those of the original-ResNet-110 and -1202 models, respectively, and 1.27% and 0.36% lower error rates than those of the pre-act-ResNet-110 and -164 models.

Fig. 5 shows the training time taken per epoch with a mini-batch size of 128 for each model. This is the time combining forward and backward time; the pre-Act-ResNet-164 model with the deepest networks takes the longest training time, and in proportion to the amount of computation when the network depths are the same, the training times taken are in the order of WR-Inception-l2, VGG, WRN-16-4, and WR-Inception.

An important point to note here is that despite the smaller amount of computations than other network models, the pre-Act-ResNet-164 model could not relatively utilize the parallel processing effect of the GPU because of its deep network. Through this, we can see that in order to accurately classify 1000 categories, a very deep thin network could have good representation power, but it is very restricted in terms of its processing speed from the perspective of the

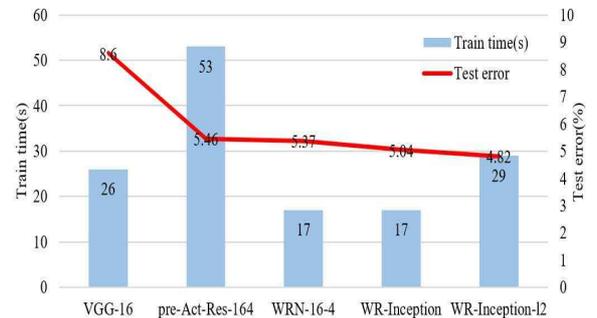
Figure 5. Train time per epoch on CIFAR-10 dataset

commercialization of deep neural networks.

## B. Transfer Learning for Object Detection

We applied the proposed WR-Inception network to the object detector SSD as feature extraction network and verified its performance on the KITTI dataset. KITTI is a dataset obtained through stereo cameras and lidar scanners in urban, rural, and highway driving environments, and has 10 categories in total, which are small cars, vans, trucks, pedestrians, sitting people, cyclists, trams, miscellaneous, and "do not care." The size of the images is 1382x512, and 7,500 images provide 40,000 object labels classified as easy, moderate, and hard, based on how much the images are occluded and truncated.

The training was conducted in a PC environment, and the test inference was on an NVIDIA Jetson TX1 board. The NVIDIA Jetson TX1 embedded board is composed of a 64-bit ARM A57 CPU, a 256-core NVIDIA Maxwell GPU at 1T-Flop/s, and 4GB of shared LPDDR4 RAM. The training method was stochastic gradient descent, and we set the mini-batch size to 32, momentum to 0.9, weight decay to 0.0005, and initial learning rate to 0.001. The learning rate decay policy was to maintain a constant learning rate, dropped by 0.1 at every 40,000th iteration. The training batch was determined by randomly selecting a 300x300 patch and warping it; the data augmentation effect of hard-negative mining was used. For equal comparison, all these training procedures were learned in the same way as the SSD [19].

We chose mean average precision (mAP), mean average recall (mAR), and processing time (in milliseconds per image) as the evaluation metrics for our experiment. mAP is an indicator that evaluates how small the detection error (wrong detection) rate is when we get precision values from each category, and sets objects with more than 50% overlap with the groundtruth box as True Positive. mAR also denotes the values of recalls from each category and gets their average, but what is different from mAP is that it evaluates how small the missed detection rate is. In the area of ADAS research, the trend is to put more emphasis on mAR than on mAP, because missed detection carries a greater risk than wrong detection in terms of safety.

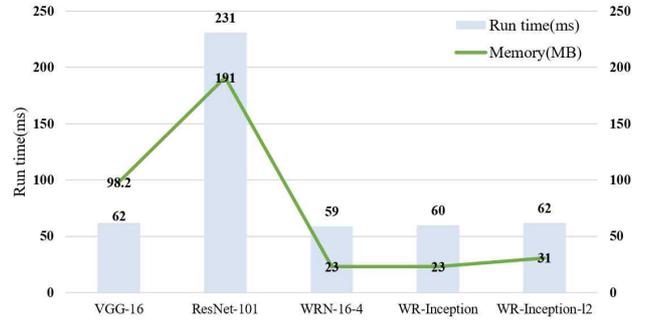

Figure 6. Comparisons of test time and weight memory on KITTI dataset

$$mAP = \text{average Precision}\left(\frac{TP}{TP + FP}\right) \quad (2)$$

$$mAR = \text{average Recall}\left(\frac{TP}{TP + FN}\right) \quad (3)$$

Table 3 is the result of KITTI object detection by different network models as the base network of SSD. When comparing a WR-Inception network to other network models, note that its mAP is higher by 4.7% to 5.3%, and mAR is higher by 4.8% to 6.14%. Through this quantitative performance enhancement, in terms of the object detection task, we were able to verify the efficiency of the proposed residual-inception unit that could see different receptive fields.

Fig. 6 displays the average test time when the network models were executed on the NVIDIA Jetson TX1 embedded board, as well as the weight memory sizes for each network model. We resized the input video to 300x300 and conducted its inference. Because the ResNet-101 model has the deepest network and many parameters, its processing time is quite a bit longer than other models. On the other hand, the proposed WR-Inception network has a shallow depth and fewer parameters than ResNet-101 but is better in terms of a higher detection rate, faster processing speed, and smaller memory size.

TABLE 3. AVERAGE PRECISION(%) & AVERAGE RECALL(%) ON KITTI VALIDATION SET

| Model | Difficulty | Car | | Pedestrian | | Cyclist | | mAP | mAR |
|---|---|---|---|---|---|---|---|---|---|
| | | AP | AR | AP | AR | AP | AR | | |
| VGG-16 | Easy | 85.00 | 98.00 | 53.00 | 71.00 | 46.00 | 75.00 | 58 | 69 |
| | Moderate | 74.00 | 75.00 | 50.00 | 56.00 | 52.00 | 71.00 | | |
| | Hard | 67.00 | 59.00 | 48.00 | 49.00 | 51.00 | 67.00 | | |
| ResNet-101 | Easy | 87.57 | 98.23 | 50.27 | 67.65 | 49.86 | 79.21 | 58.9 | 70.06 |
| | Moderate | 76.04 | 74.82 | 47.74 | 56.07 | 53.61 | 75.26 | | |
| | Hard | 68.07 | 59.54 | 45.21 | 49.17 | 51.77 | 70.55 | | |
| WRN-16-4 | Easy | 90.08 | 98.07 | 52.29 | 72.17 | 47.71 | 75.88 | 58.7 | 69.37 |
| | Moderate | 76.8 | 75.16 | 47.88 | 59.11 | 50.36 | 67.84 | | |
| | Hard | 68.5 | 59.54 | 45.3 | 52.16 | 49.38 | 64.39 | | |
| WR-Inception | Easy | 87.1 | 98.37 | 55.98 | 76 | 52.9 | 84.71 | **61.18** | **73.51** |
| | Moderate | 77.2 | 76.18 | 52.51 | 63.01 | 54.63 | 76.17 | | |
| | Hard | 68.81 | 60.13 | 48.61 | 55.41 | 52.87 | 71.58 | | |
| WR-Inception-l2 | Easy | 90.36 | 98.47 | 53.26 | 79.81 | 57.02 | 80.85 | **63.03** | **75.14** |
| | Moderate | 78.24 | 80.24 | 51.08 | 64.29 | 59.28 | 75.26 | | |
| | Hard | 71.11 | 66.54 | 49.54 | 59.44 | 57.39 | 71.37 | | |

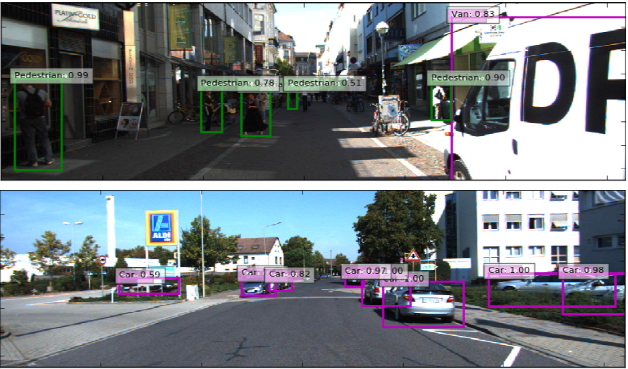
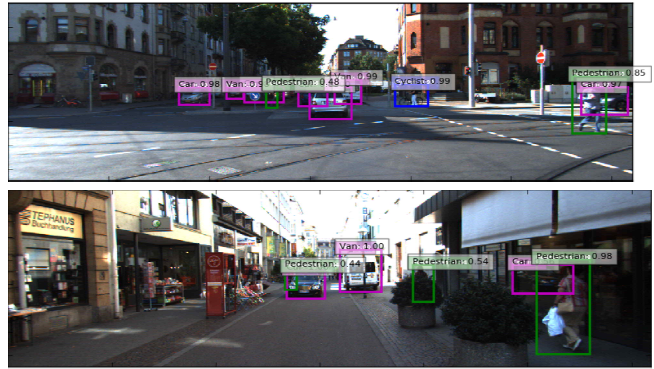
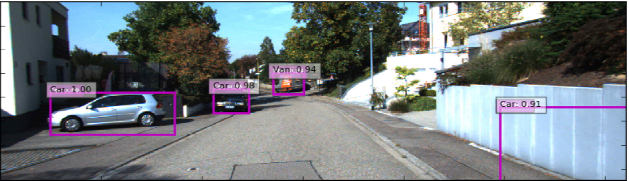

Figure 7. Results of Detections on KITTI data set

Figure 8. Results of False Positive & False Negative on KITTI dataset

Fig. 7 shows the detection result within the KITTI dataset, and Fig. 8 shows the results of false and missed detection. As shown in Fig. 8, the weakness in SSD is that it is unlikely to detect small objects. However, considering that it has very rapid processing, instead of its relatively low detection of small objects, we believe that SSD is good enough to be used commercially.

## VI. CONCLUSIONS

This paper suggested the wide-residual-inception network to overcome the limitations of the existing network models, which require a great amount of computation that limits adaptation to commercial products. We composed the overall network by using a residual connection and a residual inception unit that can see different receptive fields. When compared to the state-of-the-art ResNet-164 network (5.46%/24.33%) on the CIFAR-10/100 dataset, it shows lower error rates, 4.82%/23.12%. In addition, we transferred the proposed network to an object detector (SSD) by applying it as the feature extraction network on the KITTI dataset to verify the efficiency of the WR-Inception network. As a result, the mAP of the network was higher than that of the ResNet-101 network by 4.19%, and mAR was higher by 5.08%. Also, processing time on the NVIDIA Jetson TX1 embedded board was 62ms, which is 3.85 times faster than ResNet-101, thus proving it is capable of real-time processing, and its parameter memory was 8.3 times less than that of ResNet-101, proving it is economical and efficient in environments with limited resources, such as an embedded board or a mobile platform. Furthermore, it is expected that WR-Inception networks will be actively utilized for a variety of computer vision tasks.

As for future work, to verify the proposed WR-Inception network, it will be trained and tested on the MS-COCO dataset [34] and the PASCAL VOC dataset, which are rather general object detection tasks.

ACKNOWLEDGMENT

This work was supported by the Industrial Technology Innovation Program "10052982, Development of multi-angle front camera system for intersection AEB" funded by the Ministry of Trade, Industry & Energy (MOTIE, Korea) and partly by Institute for Information & communications Technology Promotion(IITP) grand funded by the Korea government(MSIP)(No.B0101-15-0266, Development of High Performance Visual BigData Discovery Platform for Lage-Scale Realtime Data Analysis)